****** Start of file apssamp.tex ******
\documentclass[%
 reprint,
 amsmath,amssymb,
 aps,
]{revtex4-2}

\usepackage{graphicx}
\usepackage{dcolumn}
\usepackage{bm}

\begin{document}

\title{Learn one size to infer all: Exploiting translational symmetries in delay-dynamical and spatio-temporal systems using scalable neural networks} 

\author{Mirko Goldmann}
\email{mirko@ifisc.uib-csic.es }

\author{Claudio R. Mirasso}

\author{Ingo Fischer}
\email{ingo@ifisc.uib-csic.es}

\author{Miguel C. Soriano}
\email{miguel@ifisc.uib-csic.es }
\affiliation{Instituto de F\'{i}sica Interdisciplinar y Sistemas Complejos (IFISC, UIB-CSIC), Campus Universitat de les Illes Balears E-07122, Palma de Mallorca, Spain}

\date{\today}

\begin{abstract} 
\noindent We design scalable neural networks adapted to translational symmetries in dynamical systems, capable of inferring untrained high-dimensional dynamics for different system sizes. 
We train these networks to predict the dynamics of delay-dynamical and spatio-temporal systems for a single size. 
Then, we drive the networks by their own predictions. 
We demonstrate that by scaling the size of the trained network, we can predict the complex dynamics for larger or smaller system sizes. 
Thus, the network learns from a single example and, by exploiting symmetry properties, infers entire bifurcation diagrams.
\end{abstract}
\maketitle
\section{Introduction}
Due to their many degrees of freedom and potentially multiple time scales, the prediction and analysis of complex system dynamics represent challenging tasks. 
Tackling unavailable analytical models, machine learning methods emerged \cite{lecun2015deep,Jaeger2001,gauthier2021next} that learn from data and forecast the dynamics of biological\cite{elmarakeby2021biologically,alber2019integrating}, climate\cite{dijkstra2019application}, spatio-temporal\cite{pathak2017using,pathak2018model} and other complex systems\cite{Srinivasan2022parallel,tang2020introduction,lu2018attractor}. 
More recently, incorporating symmetries is considered to guide such data-driven models towards preserving conservation laws \cite{karniadakis2021physics}, improve prediction accuracy \cite{barbosa2021symmetry} and might yield efficient representations of learned processes \cite{higgins2022symmetry}. 
Nonetheless, machine learning demands large amounts of data, and yet predictions are mainly restricted to the dynamical regime they observed during training. 
In recent works\cite{kim2021teaching, Kong2021, Kong_2021,klos2020dynamical}, parameterized neural networks are being studied for the prediction of untrained dynamics. 
Thereby, a neural network needs to be trained on several examples, often covering different dynamical regimes. 
After training, changing the parameterization renders the network the ability to generalize and predict untrained dynamics, transformations of chaotic attractors, and close-by bifurcations. 
However, in real-world applications, system parameters can often not be easily changed, and consequently, certain parameter regimes are hardly accessible. 
Here, the question arises - \textit{Is it possible to infer untrained (size-dependent) dynamical regimes of a complex system while learning from one example related to a certain system size only?}

In this manuscript, we exploit symmetries in dynamical systems by designing neural networks that exhibit i) excellent generalization properties allowing us to infer untrained dynamical properties for a wide tuning of bifurcation parameters and ii) high prediction capabilities.
Here, we design scalable neural networks that satisfy the same symmetries as the dynamical system under study. We train these networks on a single time series of either delay-dynamical or spatio-temporal systems with a fixed size. After training, notably, the size of the trained network can be scaled up or down by adding or removing neurons exploiting the translational symmetry in these systems. Without further adaptations, the scaled network generalizes from learned to untrained dynamics and enables far-reaching inferences revealing bifurcations to various dynamical regimes. In the following, we first focus on delay-dynamical systems and provide a single neural network that infers the entire bifurcation diagram for changing the delay while learning from a single example and exploiting temporal translational symmetry. Based on the analogies between delay-dynamical and spatio-temporal systems, can interpret the symmetry in delay-dynamical systems also as a quasi-spatial symmetry\cite{suppmat,arecchi1992two, giacomelli1996relationship}.  For comparison, in Section \ref{sec:st_system}, we apply our approach analogously to the real spatial translational symmetry of a Kuramoto-Sivashinsky model with periodic boundary conditions. Using a parallel network architecture we can train the network on a single spatial size and then infer untrained spatial extensions.

\begin{figure}[hb]
    \centering
    \includegraphics[width=0.375\textwidth]{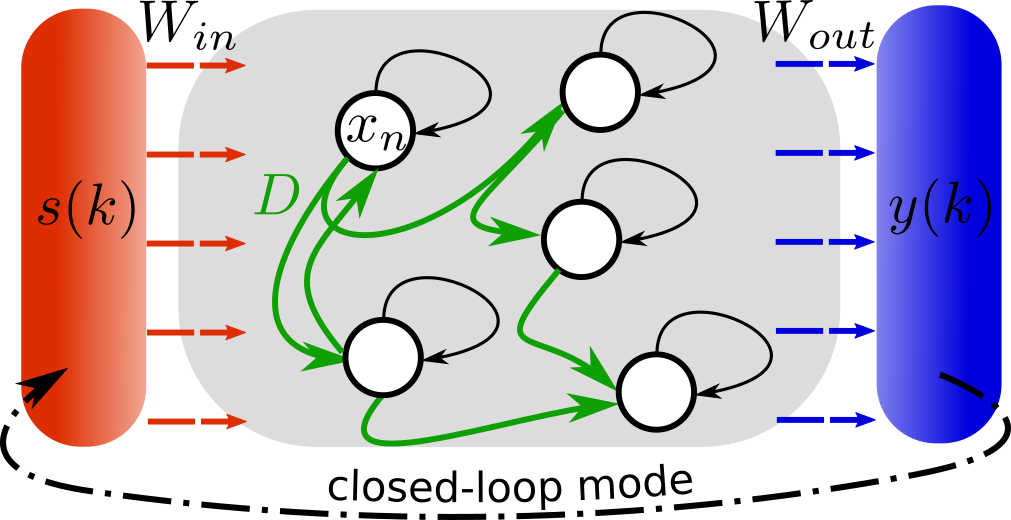}
    \caption{Scheme of the delayed echo state network. The reservoir contains delayed internode weighted connections (green), where the delay $D$ of the nodes connections can be adjusted. }
    \label{fig:scheme}
\end{figure}
\begin{figure*}[ht!]
    \centering
    \includegraphics[width=0.95\textwidth]{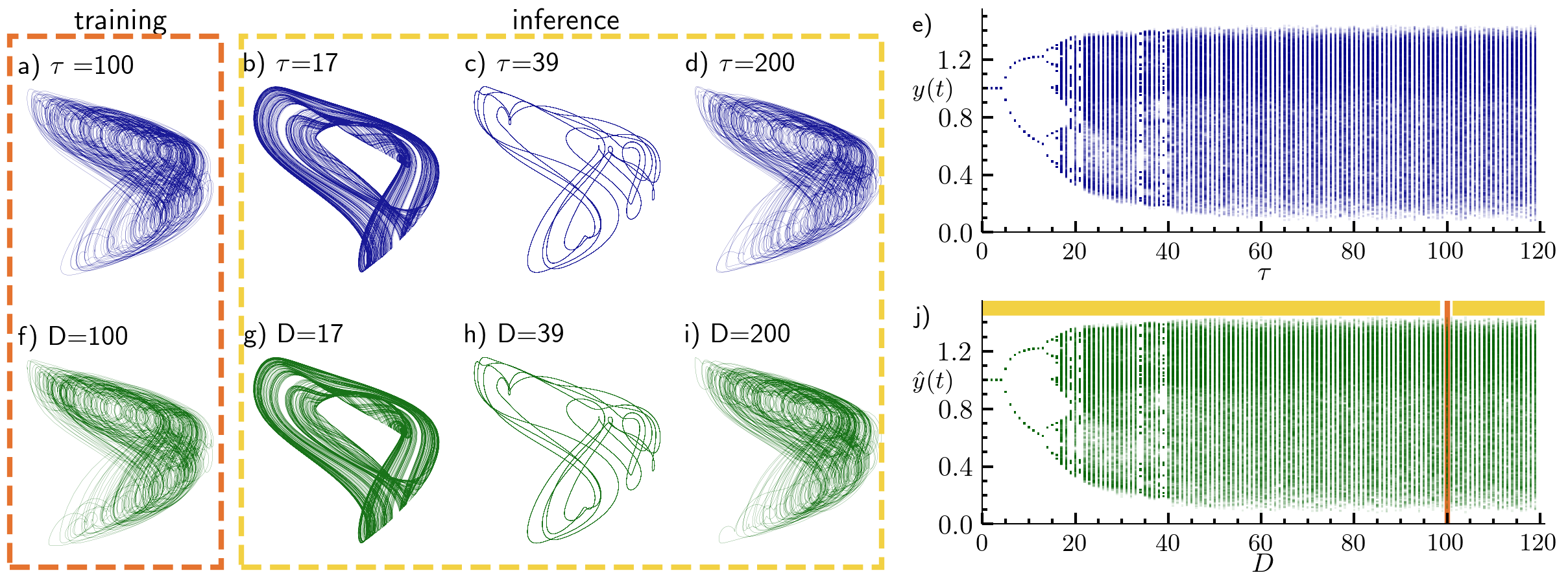}
    \caption{
    Two-dimensional projection (x-axis $y(t)$, y-axis $y(t-\tau)$) of attractors of the chaotic Mackey-Glass system for different delay lengths in a), b), c), \& d). e) Bifurcation diagram generated using the Mackey-Glass delay system. Inferred attractors by the dESN trained on a single example of the Mackey-Glass system with $\tau=100$ shown in f), g), h) \& i). j) Bifurcation diagram inferred by the dESN trained on data of a Mackey-Glass system with $\tau=100$.}
    \label{fig:bifurcDiagram}
\end{figure*}

\section{Inference of Delay System Dynamics\label{sec:delay_systems}}
Dynamical systems with delayed coupling of system variables play an important role in many real-world contexts, such as climate systems\cite{keane2017climate}, epidemiological models\cite{bauer2017chimera}, biological systems\cite{shi2017oscillatory}, control systems\cite{orosz2010traffic,stepan2009delay,zhu2019detecting} and photonic systems\cite{Appeltant2011a,larger2013,erneux2004,fischer1994high}. 
Since the evolution of delay systems relies on  a continuous history function $h(t), \ t \in [0, -\tau]$, the phase space of these systems is inﬁnite dimensional\cite{erneux2009applied}. 
Depending on the length of the delay and other system parameters, these systems either converge to fixed points, limit cycles, or evolve on chaotic attractors\cite{fischer1994high,heiligenthal2011chaos,larger2013,erneux2004}. Furthermore, with increasing delay length, these systems exhibit extensive chaotic behavior, i.e., the maximal dimension of their chaotic attractors scales linearly with the delay time \cite{LeBerre1986}.

In the following, we design echo state networks that incorporate a delay in their topology, as illustrated in Fig. \ref{fig:scheme}. 
Due to this delay, the network exhibits the same temporal translational symmetry as the delay-dynamical system it is built to predict\cite{suppmat}. 
During training, the network delays will be tuned to match those of the target system, and the readout weights are adapted to implement one-step-ahead prediction.
After learning the time evolution operator
of the delay system, the network can be scaled up or down by extending or shortening its inherent delay to infer the untrained dynamics corresponding to shorter or longer delay, respectively.
The dynamical evolution of the delayed echo state network (dESN) is governed by the following equation:
\begin{align}
\vec{x}(n{+}1) = & \alpha \vec{x}(n) {+} \beta \tanh(\mathbf{W}\vec{x}(n{-}D) + \gamma \mathbf{W}_{in}s(n) +\mathbf{W}_b)
\label{eq:dESN}
\end{align}
where $\vec{x}(n)\in\mathbb{R}^K$, $K$ is the network network size, $\alpha$ is the leak term, $\beta$ is a feedback gain, $\gamma$ is the input gain, and $D$ is the delay length. 
The total size of the network scales linearly with $D$, as all previous states $x(n)$ up to $x(n-D)$ must be stored. Thus, the delay can be seen as a signal through a cue of $D$ hidden linear neurons along each delayed connection. However, the read-out dimension of the network is independent of the delay length since $\mathbf{W}_{out}\in \mathbb{R}^K$. 
The randomly drawn matrix $\mathbf{W}_{in}\in\mathbb{R}^{K}$ gives the connection between the input $s(n)\in\mathbb{R}$ and the network, while $\mathbf{W}_{b}\in\mathbb{R}^{K}$ gives a random bias to each node. 
The delayed connections are weighted via $\mathbf{W}\in\mathbb{R}^{K\times K}$, the elements of which are randomly drawn from a uniform distribution $\mathcal{U}[-1,1]$, with a sparsity of $1.5\%$. 

\begin{figure}[ht!]
    \centering
    \includegraphics[width=0.45\textwidth]{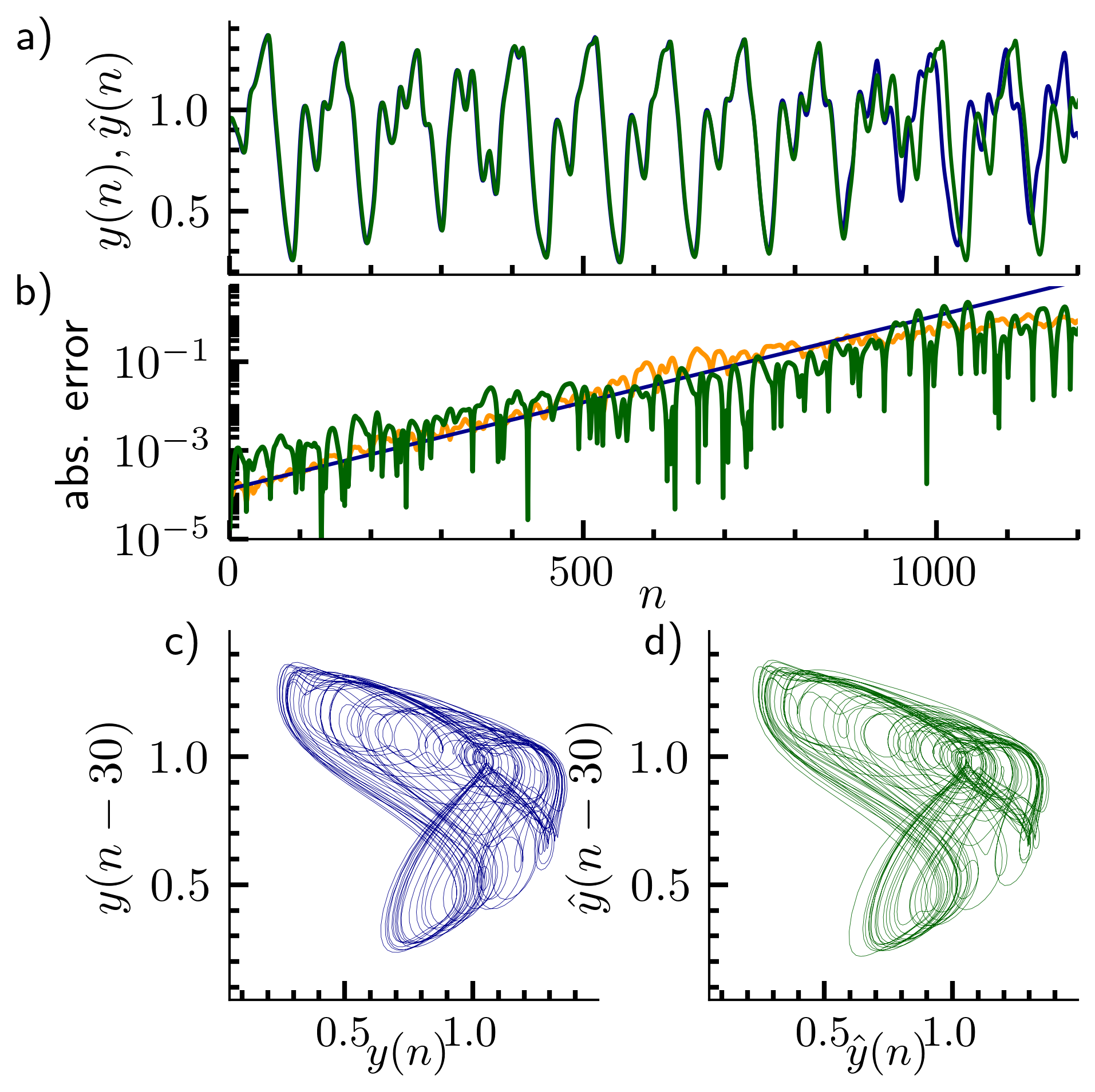}
    \caption{
    a) Time series of the original Mackey Glass system (blue) and the autonomously continued (green) chaotic attractor using a dESN. b) Divergence rate of the chaotic system and the time series generated by the dESN (green) as shown in a), the orange line indicates the average divergence rate of 20 different dESNs initialized at 20 different trajectories of the chaotic Mackey-Glass system. The blue line indicates the divergence rate related to the largest Lyapunov exponent $\lambda=0.009$ of the Mackey-Glass system with $\tau=30$. In c) and d), two-dimensional projection of the chaotic attractor of the Mackey-Glass system with a delay of $\tau=30$ and the dESN prediction with $D=30$, respectively.}
    \label{fig:MGoverview_30}
\end{figure}

The training data set used only contains a time series of a delay system with a single fixed delay. 
Here, we consider inferring the dynamics of a Mackey-Glass delay system\cite{mackey1977oscillation}, however, the same holds for other delay systems as we demonstrate in Appendix \ref{app:ikeda} by performing a similar study for an Ikeda-type delay system. 
The dynamics of the Mackey-Glass (MG) delay system is given by the following equation:
\begin{equation}
    \dot{s}(t)= - \frac{s(t)}{T_0} + \dfrac{0.2s(t-\tau)}{1 + s(t-\tau)^{10}}.
    \label{eq:mg}
\end{equation}
We set the characteristic relaxation time $T_0=10$ and the system delay $\tau=100$ and generate a time series $s(t)$ that is sampled with $\Delta t=1$ to obtain the training data set $s(n)$. For these parameters, the system evolves along a chaotic attractor as depicted in Fig. \ref{fig:MGoverview_30}.
\begin{figure*}[ht!] 
    \centering
    \includegraphics[width=0.95\textwidth]{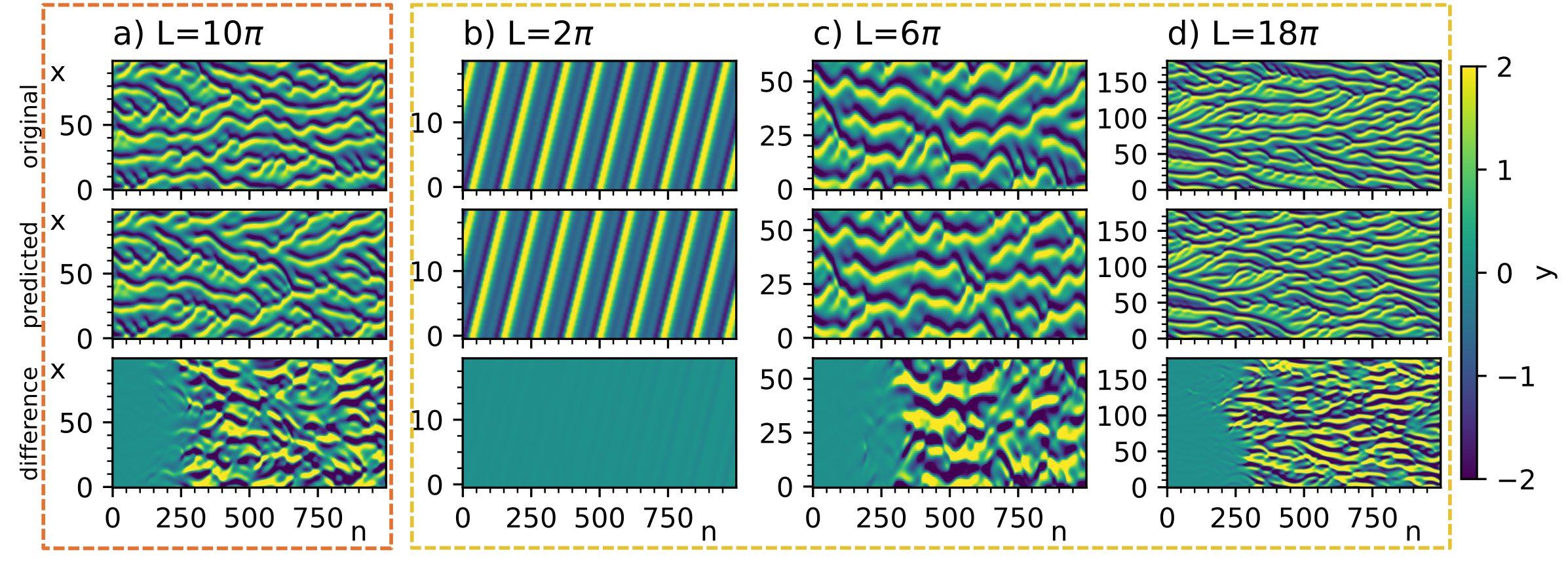}
    \caption{The dynamics of the spatio-temporal Kuramoto-Sivashinsky system $y(x,n)$ (top row), the predicted dynamics from a parallel network architecture $\hat{y}(x,n)$ (mid row), and their difference $y(x,n)-\hat{y}(x,n)$(bottom row). The parallel networks are trained with data from the Kuramoto-Sivashinsky system of spatial extension $L=10\pi$ (red box), by adapting the network architecture it generates the dynamics also for smaller and larger spatial extensions (yellow box). For comparison both systems were initialized with the same initial conditions.} 
    \label{fig:ks_pde}
\end{figure*}
Before training the reservoir on the MG time series, the delay $D$ of the dESN must be adjusted to the delay of the target system. We find the optimal performance when we match the delay of the reservoir connections directly with the delay of the MG system, i.e. by setting $D=\tau/\Delta t = 100$. In most real-world systems, however, the delay is not known a priori and needs to be extracted from the data. As we show in the Appendix \ref{app:delay_estimation}, we can optimize the reservoir delay $D$ by determining the best performance in a one-step ahead prediction that reveals the delay underlying the training data. 
All other hyperparameters of the dESN are optimized using Bayesian optimization and are given in table \ref{table:hyperparameter}.
\begin{table}[h]
    \centering
    \begin{tabular}{|l|l|l|}
        \hline
        parameter       & symbol                  & value \\ \hline \hline
        network size    & $K$                     & 1000  \\
        initial steps   & $N_{init}$              & 5000  \\
        training steps  & $N_{train}$             & 25000  \\
        feedback gain   & $\beta$                 & 0.176      \\
        input gain      & $\gamma$                &     1.24  \\
        spectral radius & $\rho$                  &      0.84 \\
        leak rate       & $\alpha$                &      0.75\\ \hline
    \end{tabular}
    \caption{List of parameters used for learning the Mackey-Glass system.}
    \label{table:hyperparameter}
\end{table}

During training, the dESN is driven by the input $s(n)$ with added white noise. The addition of noise during training was shown to improve the stability of the reservoir in closed loop mode\cite{estebanez2019constructive}. The network's output layer is optimized using linear regression to predict the next step in the training time series. Once the output weights $W_{out}$ are computed, the reservoir is decoupled from the input $s(n)$ and the prediction of the reservoir $\hat{y}(n)=\mathbf{W}_{out} \vec{x}(n)$ is returned as the new input. Consequently, the reservoir evolves autonomously in closed-loop mode\cite{JAE04, Antonik2018, Haluszczynski2019}. Reservoir computing \cite{Jaeger2001, MAA02,nakajima2021reservoir,lukovsevivcius2012reservoir} is applied here, as it offers fast and efficient training; nevertheless, we expect that other training methods yield comparable results. 

In closed-loop mode, the reservoir evolves along the attractor on which it was trained, as depicted in Figs. \ref{fig:bifurcDiagram} a) and f). When resetting the delay $D$ of the dESN connections, we do not need any additional data for initialization, no further retraining, and the output weights can be kept fixed.
As shown in Figs. \ref{fig:bifurcDiagram} g) to i), the dESN generates high-dimensional chaotic attractors at $D=17, \ 200$ and a limit cycle at $D=39$. These inferred dynamical states are similar to those of the original MG system with delay $\tau=17, \ 39$ and $200$ shown in Figs. \ref{fig:bifurcDiagram} b) to d). Consequently, it is sufficient that the dESN learned from a single time series of the MG system with delay $\tau=100$, to infer the dynamics for other delay lengths. By scanning the delay of the dESN from $D=1-120$ we generate a bifurcation diagram of the inferred dynamics as shown in Fig. \ref{fig:bifurcDiagram} j). Depending on $D$, the dESN exhibits chaotic behavior, transitions to intermittent limit cycles, for example, at $D=34 \ \&  \ 39$, period-doubling bifurcations at $D=14 \ \& \ 16$, and transitions to a stable fixed point at $D=5$. The observed transitions in the dESN coincide with those found for the original MG system shown in Fig. \ref{fig:bifurcDiagram} e). Furthermore, in Appendix \ref{sec:coexistence} we show that the dESN can infer multistabilities in the delay range $\tau,D=70-80$. Röhm et al.\cite{rohm2021modelfree} recently presented that reservoir computing is capable of predicting multistabilities in a dynamical system even if the reservoir is trained only around one of the attractors. Here, our dESN infers infers coexisting attractors, while it was trained in a dynamical regime where there was no multistability. 

In Fig. \ref{fig:bifurcDiagram} we show that the dESN trained at the MG system with $\tau=100$ can infer dynamics of the MG system with other delay lengths. In the following, we further underline how precise the dESN infers the dynamics even far from the training example. 
Therefore, in Fig. \ref{fig:MGoverview_30}, we show the inference of the dESN with $D=30$ compared to the original dynamics. 
We divide the prediction capabilities into two regimes, which we term weather and climate\cite{estebanez2019constructive}. 
Here, the weather regime refers to the short-term behavior, when going from externally driven operation to autonomous operation. 
As shown in Fig.\ref{fig:MGoverview_30} a), the dESN can precisely continue the trajectory of a MG system with $\tau=30$ for around $1000$ steps, which corresponds to approximately nine Lyapunov times. 
Due to the chaotic behavior of the MG system with $\tau=30$, two closely initialized trajectories diverge in time, where the divergence rate is given by the largest Lyapunov exponent. 
As depicted in Fig. \ref{fig:MGoverview_30} b), the divergence rate of the predicted trajectory (green line) is similar to the largest Lyapunov exponent of the MG system (blue line). 
Furthermore, the performance that the dESN provides is robust against randomization. We find similar results (orange line) for 20 different dESNs with varied connection matrices and training data. 
The climate regime describes the long-term behavior of a dynamic system, e.g. the evolution along the chaotic attractor. 
In Fig. \ref{fig:MGoverview_30} d), we show the predicted trajectory after it diverges from the initialization trajectory. 
The dESN reproduces the climate of the delay system by inferring the chaotic attractor of the MG system with $\tau=30$ as shown in Fig. \ref{fig:MGoverview_30} c) and is also the case for other delays $D$ shown in Fig. \ref{fig:bifurcDiagram}. 
Accordingly, training the dESN only on a single example is sufficient to precisely infer the dynamical properties of the MG system for various delay lengths.


As presented in Figs. \ref{fig:bifurcDiagram} and \ref{fig:MGoverview_30}, training the dESN with data from the MG system with a delay of $\tau=100$ enables inferring untrained dynamics related to shorter and even longer delays. A delay of $\tau=100$ places the system in the long delay limit, where the delay is much longer than the characteristic relaxation time $T_0$ in Eq. \ref{eq:mg}. Here, learning benefits from a clear time-scale separation between response and delay time. We find that learning in the short delay regime enables inferring towards relatively shorter delays, whereas going towards larger delays, the inference quality deteriorates. For a more detailed discussion, we refer to Appendix B.


\section{Inference of Spatio Temporal system dynamics \label{sec:st_system}}

Taking into account the close relationship of delay and spatio-temporal systems \cite{arecchi1992two,giacomelli1996relationship,suppmat}, we extend our approach to the spatial translational symmetry of a homogeneous Kuramoto-Sivashinsky (KS) model with periodic boundary conditions. By analogy, we design reservoirs that can infer attractors for different system sizes when trained for a single size only. The dynamics of the KS model are governed by:
\begin{equation}
    y_t = - yy_x - y_{xx} - y_{xxxx},
\end{equation}
where $y(x,t)$ is a scalar field. We further consider periodic boundary conditions in the interval $[0, L)$. For the training, we generate a data set using a spatial extension of $L=10\pi$. The generated data is sampled every $\Delta t = 0.25$ in the temporal domain and contains $Q=100$ equidistant samples in the spatial domain leading to $y(n)\in \mathbb{R}^{100}$. To predict the spatio-temporal evolution of the KS system, we construct a parallel network architecture as described by Pathak et al.\cite{pathak2018model} and train this architecture to perform a one-step-ahead prediction. The parallel architecture contains $G=10$ subnetworks, each containing $K=1000$ neurons, respectively. The evolution of the $g$-th subnetwork can be described by the following equation: 
\begin{equation}
    x^g(n+1) = \tanh(W x^g(n) + \gamma W_{in} u^g(n) + 0.2W_b ),
\end{equation}
where $W\in\mathbb{R}^{N \times N}$ is a randomly drawn adjacency matrix with spectral radius $\rho=1.3$, $W_{in}\in\mathbb{R}^{N\times M}$ are randomly drawn input weights, the input gain $\gamma=0.001$, and $W_b\in\mathbb{R}^{N}$ is a random bias. The input $u^g(n)$ of the subnetworks is generated by dividing the data from the spatial domain into $G$ sections of the same size $Q/G=10$ (spatial size of $\pi$). Each of the $G$ subnetworks receives the 10 inputs of one particular section and additionally the 3 closest inputs from both neighboring sections, leading to an input dimension of $M=16$. Due to the spatial translational symmetry of the homogeneous KS system along the spatial domain the subnetworks adjacency matrix and hyperparameters can be chosen identical. Similar approaches are mentioned in \cite{pathak2018model, barbosa2022learning}. During training, each subnetwork is trained to perform a one-step-ahead prediction of the spatial domain section to which it corresponds, resulting in 10 outputs per network computed using the output weight matrix $W_{out}\in\mathbb{R}^{10 \times 1000}$. Again, due to the translational symmetry the output weights sets can be shared between the subnetworks. After the training phase, we close the loop by feeding the predicted state back to the reservoir. As shown in Figs. \ref{fig:ks_pde} a) and f), parallel reservoirs autonomously predict the weather and climate of the chaotic KS system with $L=10\pi$. 

To infer spatial extensions $L$ that are different from the one used during training, we take further advantage of the spatial translation symmetry and similar to changing the delay in the dESN, we adapt the topology of the reservoir. Therefore, either a subnetwork is removed from the parallel architecture or a copy of a subnetwork is inserted into the architecture. By varing the number of subnetworks $G$ in the architecture the output dimension alters in steps of $10$ which effectively increases the predicted spatial extensions.  Due to the spatial length trained on and the used number of subnetworks, we can vary the spatial extension of the predicted system in units of $\pi$. In Fig. \ref{fig:ks_pde}, we present the original KS dynamics (first row), the predicted dynamics of the reservoir (middle row), and the difference between both (bottom row) for different spatial extensions, respectively. Thereof, we initialized both using the same initial conditions. In analogy to the results for the MG delay-dynamical system, we observe that the once-trained reservoir is able to infer the untrained dynamics at significantly shorter and larger spatial extensions. Therefore, the parallel network architecture takes advantage of the spatial translational invariance of the homogeneous KS model with periodic boundary conditions.

\section{Conclusion}
We demonstrated that exploiting symmetries of dynamical systems by designing neural networks obeying these symmetries improves their prediction ability and further enable far-reaching inference. We particularly showed this for translational symmetry in delay-dynamical and spatio-temporal systems.
Transformation of the trained networks along this symmetry enabled learning from a single example to infer an entire bifurcation diagram. 
Thus, we obtain minimal requirements for training data and gain a single model that can infer a wide variety of dynamical behaviors. 
This represents a very efficient use of resources, such as the required network size, training data, and energy. 
Therefore, it is also a step in the direction of more sustainable machine learning.  
In addition, the provided method might be used to analyze real-world systems for which certain parameter settings might not be accessible. 
Recently, Tegmark et al.\cite{liu2022machine} presented a machine learning method to manifest hidden symmetries in physical systems. 
The discovery of hidden symmetries and the use of our approach to exploit them for far-reaching inferences while learning from single examples could give rise to powerful predictive models for a variety of high-dimensional systems, including complex networks\cite{Srinivasan2022parallel}. 

\section{appendix}
\appendix
\section{Delay estimation from time series using delayed echo state networks\label{app:delay_estimation}}

\begin{figure}[ht!]
    \centering
    \includegraphics[width=0.45\textwidth]{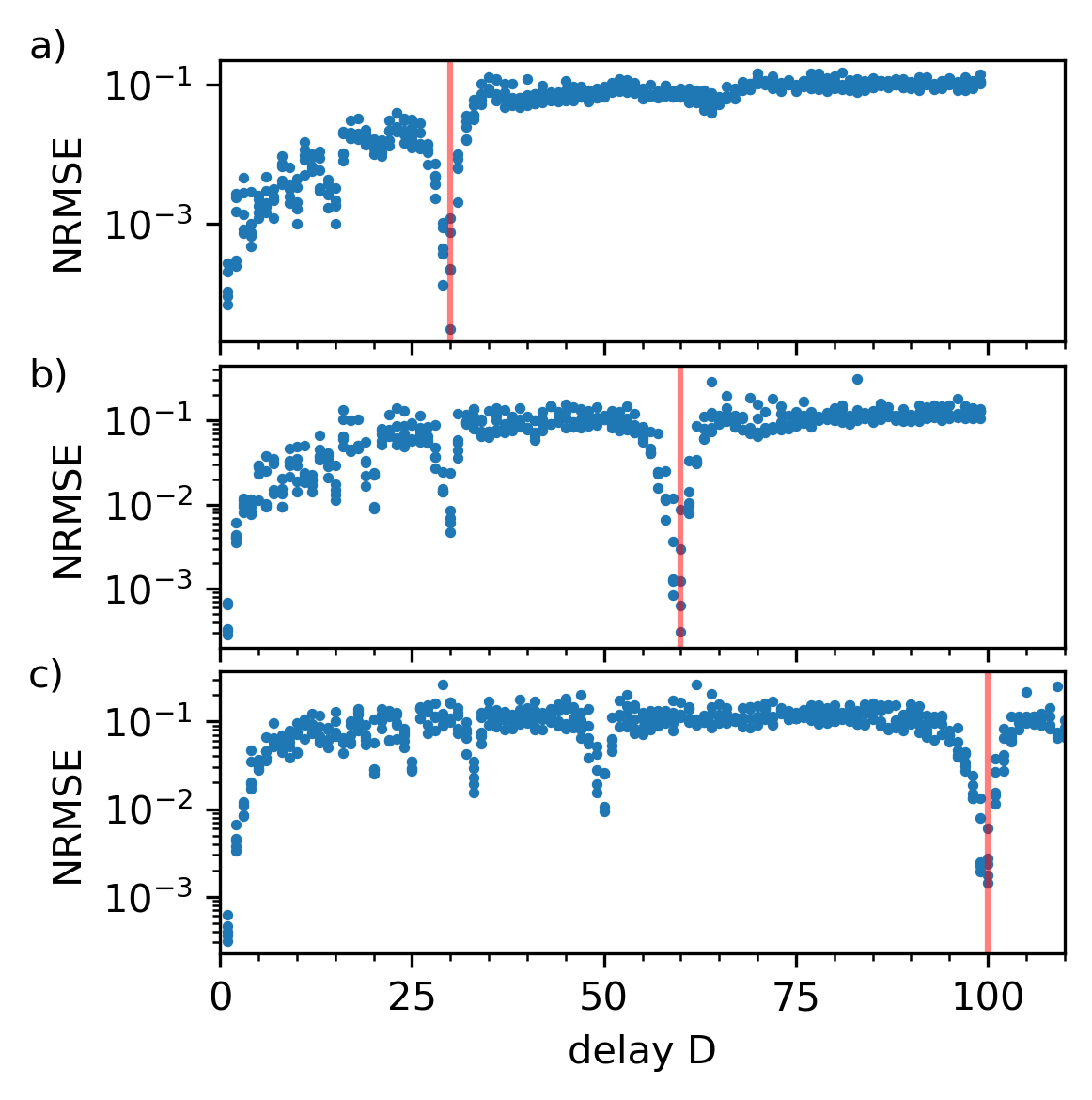}
    \caption{Estimating the delay a from time series of the chaotic Mackey-Glass system for three different delay lengths $\tau$. The red line marks the delay $\tau$ that underlies the data set a) $\tau=30$, b) $\tau=60$ and c) $\tau=100$. The blue dots mark the one-step-ahead prediction accuracy of dESNs with different delays $D$ (x-axis) and different sets of hyperparameters. The accuracy is determined by the NRMSE (normalized root mean square error) of the one step ahead prediction.}
    \label{fig:findDelay}
\end{figure}

As mentioned in section \ref{sec:delay_systems}, during training, we match the delay of the delayed echo state network (dESN) with the delay of the Mackey-Glass system to enhance the prediction abilities of the dESN. As the Mackey-Glass time series is sampled with $\Delta t=1$ this leads to $D=\tau$. In many real-world systems, however, the underlying delay is not known a priori and needs to be determined from a sampled time series of the dynamical system under study. There exist several methods to estimate delays from time series based on, e.g., autocorrelation function, delayed mutual information, local linear fitting in a low-dimensional subspace\cite{hegger1998identifying}, and even deep learning-based methods \cite{Gao2021delayestimation}. In the following, we show how to estimate the delay that underlies a time series using the delayed echo state network (dESN) given in Eq. \ref{eq:dESN}. Thus, we seek for high one-step-ahead prediction accuracy of dESNs by scanning the delay $D$ of the dESN and optimizing the reservoir hyperparameters using Bayesian optimization respectively. In Fig. \ref{fig:findDelay}, we show the results of this scan for three different delays $\tau=30,60,100$ of the Mackey-Glass system that generated the training data. The delay scan of the dESN reveals an increased performance (reduced NRMSE) when its delay is in resonance or equal to the delay $D = \tau$ of the Mackey-Glass system. As it can be seen in Fig. \ref{fig:findDelay}, identifying these resonances in turn allows to extract the delay that underlies the data. There can be relative offsets of $\pm1$ in the optimal estimated delay, which here is the smallest possible offset related to the sampling of the Mackey-Glass time series used here ($\Delta t=1$). In this manner, the optimal setting of the dESN delay might depend on the frequency used for sampling the real-world system. 

\section{Dependence of inference capabilities on the trained chaotic attractor}

\begin{figure}[htb]
    \centering
    \includegraphics[width=0.475\textwidth]{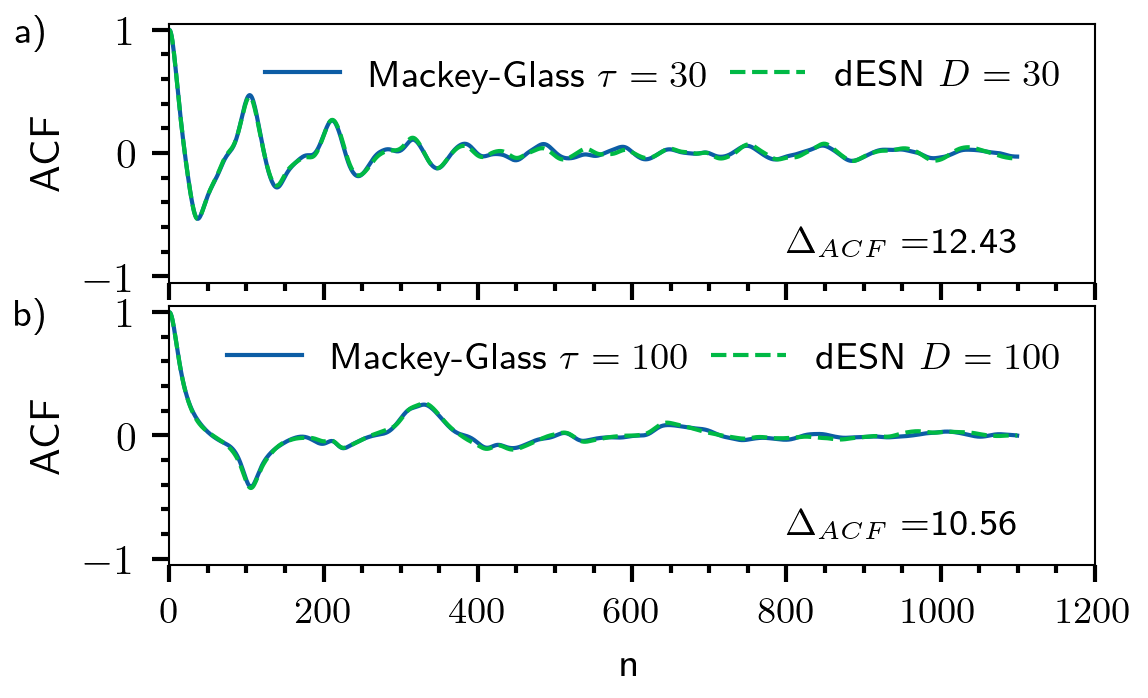}
    \caption{Comparison of the autocorrelation function ACF of the Mackey-Glass system (blue) with a) $\tau=30$ and b) $\tau=100$ and the predicted time series using the dESN (green). The dESN used in this plot was trained on a chaotic Mackey-Glass system with $\tau=100$.}
    \label{fig:acf_example}
\end{figure}

\begin{figure}[hbt]
    \centering
    \includegraphics[width=0.45\textwidth]{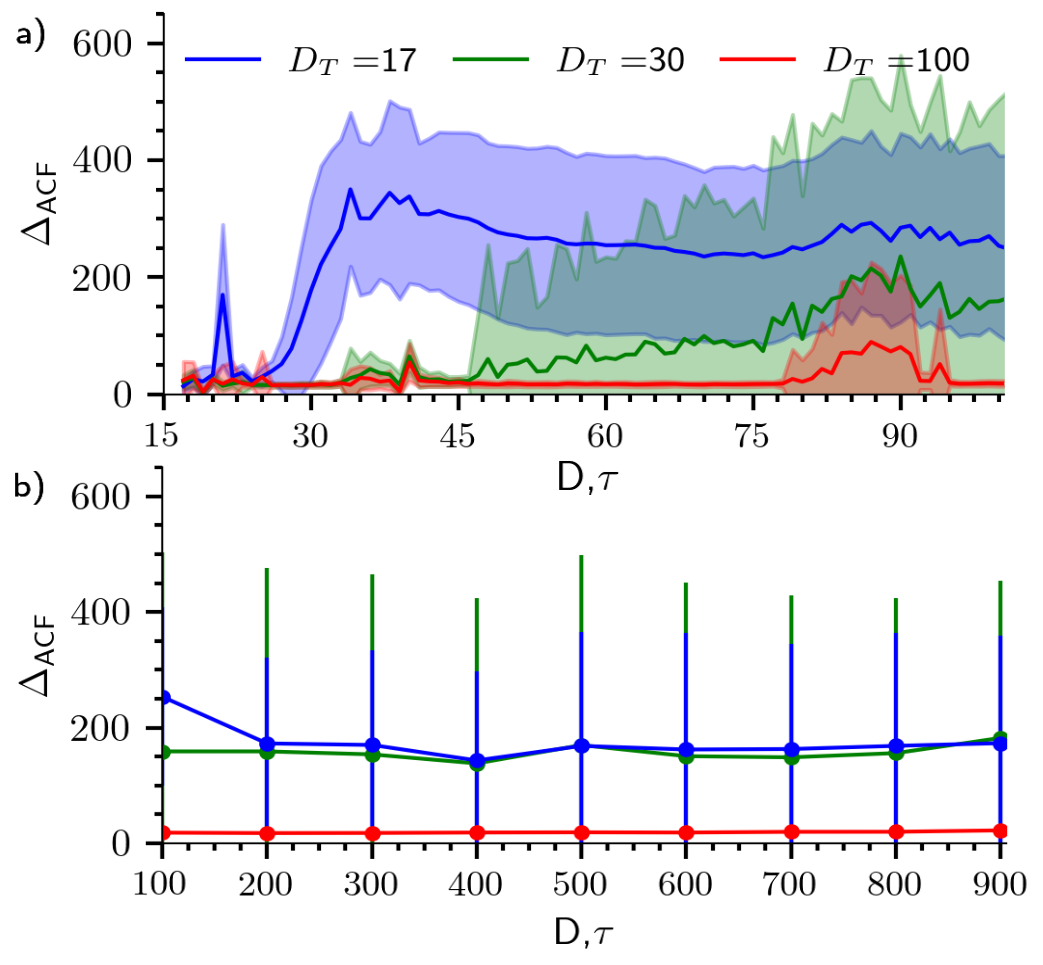}
    \caption{Absolute difference of the original and the predicted autocorrelation $\Delta_{ACF}$ of the chaotic attractors a) in the range $\tau,D\in[17,100]$ and b) $\tau,D\in[100,900]$. The colors refer to different delays of the Mackey-Glass system used in the training of the dESN, $\tau=17$ (blue), $\tau=30$ (green) and $\tau=100$ (red). Solid lines indicate the average over 100 random seeds used to generate the dESN and initialize the Mackey-Glass system that generated the training data. The shaded areas indicate the standard deviation, respectively.}
    \label{fig:acf_long_delays}
\end{figure}

In section \ref{sec:delay_systems}, we show that the dESN with delay $D=100$ trained only on the Mackey-Glass time series with $\tau=100$ can predict the entire bifurcation diagram of the Mackey-Glass system by scaling its size (parameter $D$) after training. This means that the scalable dESN is able to infer dynamics of much shorter but also longer delays while being trained on data of a single delay length only. In the following, we evaluate how these prediction capabilities depend on the properties of the chaotic attractor used for the training of the dESN. To do this, we use the absolute difference between the autocorrelation, denoted $\Delta_{ACF}$, of the Mackey-Glass attractor and the attractor predicted by the respective dESN to quantify the quality of the prediction. Here, we computed the autocorrelation (ACF) of the original and predicted time series for 1100 steps into the past, as shown in Fig. \ref{fig:acf_example}. Subsequently, the absolute difference $\Delta_{ACF}$ between the original and predicted ACF is calculated and summed in the range $n\in[0,1100]$.
For comparison, we train three networks where $D_T=17,\ 30 \ \& \ 100$ defines the delay during training, respectively, and the corresponding training data set is given by the Mackey-Glass time series with similar corresponding delay $\tau=17,\ 30 \ \& \ 100$. In Fig. \ref{fig:acf_long_delays} we show the autocorrelation difference $\Delta_{ACF}$ for two ranges: a) $\tau, D=17-100 $ and b) $\tau,D = 100-900$. While all trained dESN can predict towards smaller delays, the predicted attractors from the dESNs trained at $D_T=17$ (blue) and $D_T=30$ (green) start to deviate from the original Mackey-Glass attractor at a delay $D>30$ and $D>50$, respectively. In contrast, the one trained at $D=100$ (red) does not show an increase in the difference between the predicted and original autocorrelation function $\Delta_{ACF}$ even if the delay $D$ of the dESN becomes much larger than the one used during training (deviations in the range $\tau, D = 75-95$ are caused by multistability, see appendix \ref{sec:coexistence}). These results illustrate that the dESN $D_T=100$ trained at $\tau=100$ has the highest capabilities and can predict even attractors with much longer delays.

We relate this ability of the dESN with $D_T=100$ to the properties of the dynamics observed during training. As mentioned in the section \ref{sec:delay_systems}, a delay of $\tau=100$ sets the Mackey-Glass system in the long delay limit, where a clear separation of the local and the delayed dynamics appear. This can be indicated by using the autocorrelation function of the time series, as shown in Fig. \ref{fig:acf_example}, wherein panel a) a short delay $\tau=30$ does not show a clear separation between delay and local dynamics. In contrast, in panel b) of Fig. \ref{fig:acf_example}, one can observe this separation showing a decay of local dynamics until $n=50$ and the delayed dynamics indicated by the peak around the delay time $n=100$. 

In conclusion, exploiting translational symmetries enables one to infer untrained dynamics while learning from a chaotic time series related to a single system size only. The prediction ability of the trained scalable neural network further depends on the (size-dependent) dynamical regimes on which it was trained. Thereby, the prediction ability of the dESN is the strongest if the network is trained in the long delay limit of delay-dynamical systems. 

\section{Inferring coexisting limit cycle and chaotic attractor \label{sec:coexistence}}

\begin{figure}[h]
    \centering
    \includegraphics[width=0.45\textwidth]{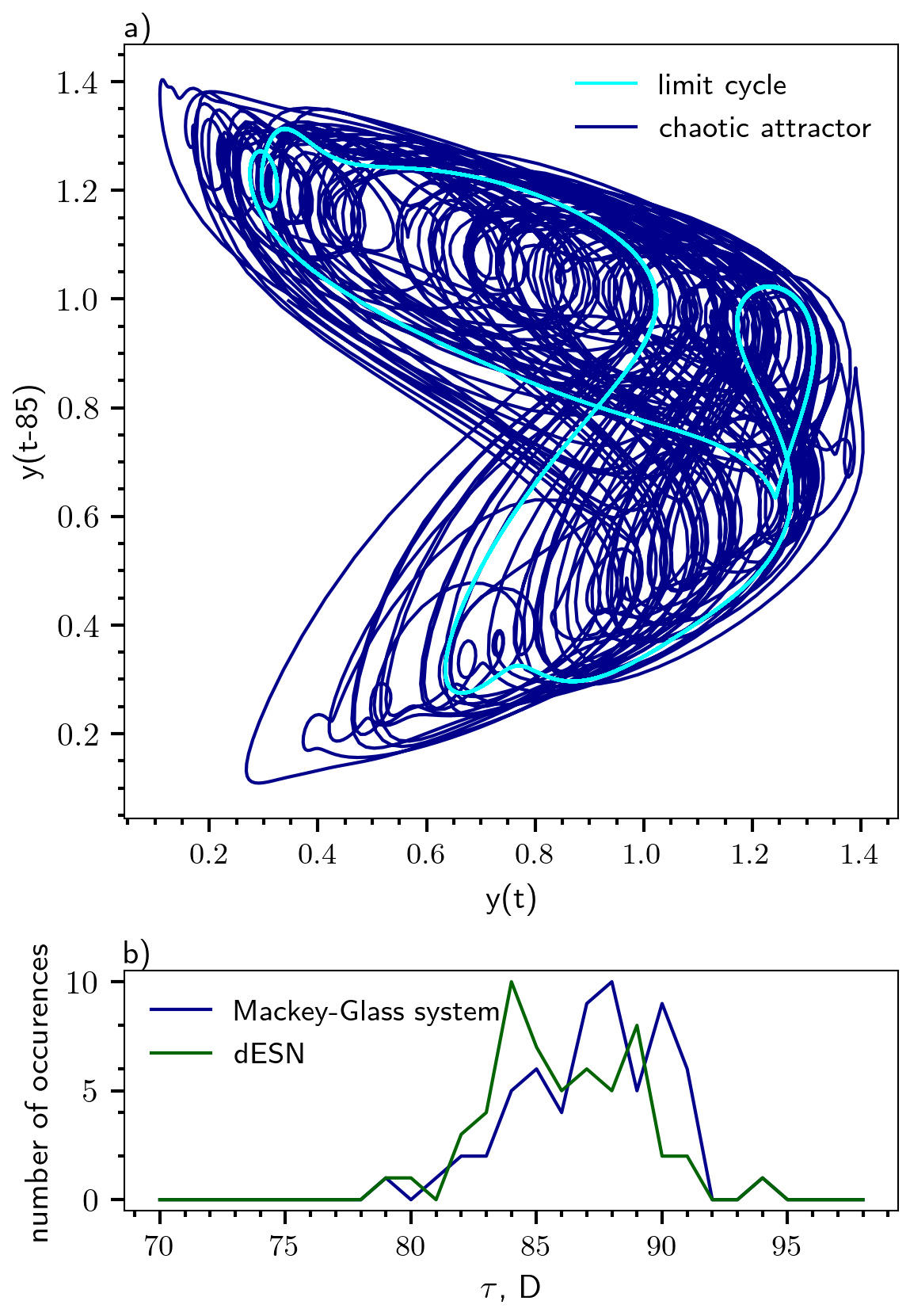}
    \caption{a) 2D representation of the coexisting limit cycle and chaotic attractor generated by the Mackey-Glass system at a delay of $\tau=85$. b) Number of occurrences of a limit cycle in the delay range $\tau,D\in[70,100]$ for 100 different initial conditions. The blue colored line indicates the occurrences in differently initialized Mackey-Glass systems, while green indicates occurrences in the differently initialized dESN systems. The dESN is trained at $D=100$ on a Mackey-Glass system with $\tau=100$ and afterwards the delay D was scaled down.}
    \label{fig:coexistence}
\end{figure}

As mentioned in the manuscript, the Mackey-Glass system exhibits a multistability leading to coexistence of a limit cycle and a chaotic attractor in the region $\tau\in[79,95]$. In Fig. \ref{fig:coexistence} a), we show an overlay of the limit cycle and the chaotic attractor for a delay of $\tau=85$. During the computation of the bifurcation diagram, depending on their initial conditions, the trajectory of the Mackey-Glass system and the dESN converge either to one or the other of the attractors. This effect causes deviations in the computation of the autocorrelation comparison shown in Fig. \ref{fig:acf_long_delays}. 
In Fig. \ref{fig:coexistence} b), we quantify the frequency with which the Mackey-Glass system and dESN end up in the limit cycle, respectively. Therefore, we generate time series of both systems starting from 100 different initial conditions. Using the autocorrelation function of the time series generated, we discriminate if the observed attractor is either chaotic or a limit cycle. As shown in Fig. \ref{fig:coexistence} b), both the Mackey-Glass system and the dESN exhibit comparable probabilities of ending in the limit cycle. It is worth mentioning that the dESN was trained with data of a Mackey-Glass system with $\tau=100$ where there exists no multistability. The prediction of attractors in a multistable system that was not part of the training using RC was shown recently by Röhm et al.\cite{rohm2021modelfree}. The fact that the dESN can predict these multistabilities even if it is trained on a Mackey-Glass system with a different delay illustrates how strong the generalization ability of the dESN is.

\section{Inferring Ikeda delay system dynamics \label{app:ikeda}}

\begin{figure*}[ht!]
    \centering
    \includegraphics[width=0.95\textwidth]{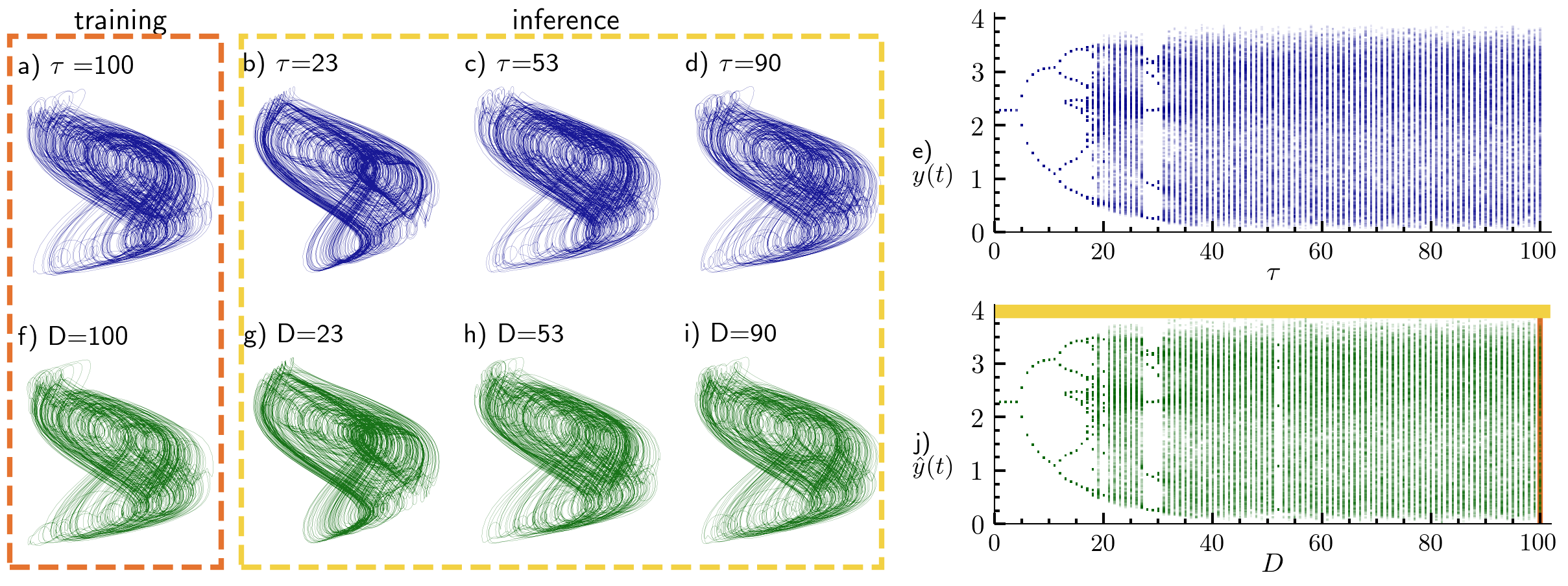}
    \caption{Two-dimensional projection (x-axis $y(t)$, y-axis $y(t-\tau)$) of attractors of the chaotic Ikeda system for different delay lengths in a), b), c), \& d). e) Bifurcation diagram generated using the Ikeda delay system. Inferred attractors by the dESN trained on a single example of the Ikeda system with $\tau=100$ shown in f), g), h) \& i). j) Bifurcation diagram inferred by the dESN trained on data of a Ikeda system with $\tau=100$.}
    \label{fig:ikedaBif}
\end{figure*}

In the main text, we show the autonomous continuation of a Mackey-Glass delay system for short, medium, and long delays and the prediction of the bifurcation diagram for arbitrarily long delays. However, the Mackey-Glass system is only a single example of a delay dynamical system, and there exists a variety of such delay systems featuring other nonlinearities. In the following, we show that the proposed method of using a dESN to predict the bifurcation diagram is not only restricted to the Mackey-Glass system, but can be applied to other delay systems. Another well-known delay system, called the Ikeda delay system, was investigated in the field of optics because it can be generated using a Mach-Zehnder interferometer and a cavity\cite{larger2013}. Here, we attempt to learn and infer the dynamics of an Ikeda-type delay system with a sine-square nonlinearity:
\begin{equation}
\dot{s}(t) =  -1/T_0 s(t) + \beta sin^2(s(t-\tau)).
\label{eq:ikeda}
\end{equation}
The state of the delay system is given by $s(t)$, the delay is indicated by $\tau$, the characteristics relaxation time is fixed at $T_0=10$ and the feedback gain is $\beta=0.4$ in the following. The data set is generated by integrating the delay differential equation in Eq. (\ref{eq:ikeda}) and the time series is sampled with $\Delta t=1$. Due to the delay, this Ikeda system can evolve in different dynamical regimes including chaos \cite{larger2013,erneux2004}. In the following, we will train a dESN to continue the time series of the Ikeda delay system with a delay of $\tau=100$. Afterward, the trained dESN will be used to infer the entire bifurcation of the Ikeda system.

As in the case of the Mackey-Glass system, the delay $D$ of the dESN and the delay $\tau$ of the Ikeda system to be learned are matched. Afterwards, the other hyperparameters of the dESN are optimized using a Bayesian optimization approach. The found optimal parameters are given in Tab. \ref{tab:ikeda}. The optimal leak rate to predict the Ikeda delay system is found at $\alpha=0$, which means that the leak term in the equation of the dESN can be neglected. 
\begin{table}[h]
\centering
\begin{tabular}{|l|l|l|}
\hline
parameter       & symbol                  & value   \\ \hline \hline
network size    & $K$                     & 1000    \\
initial steps   & $N_{init}$              & 1000    \\
training steps  & $N_{train}$             & 20000   \\
feedback gain   & $\beta$                 & 0.1     \\
input gain      & $\gamma$                & 1.71429 \\
spectral radius & $\rho$                  & 0.6975  \\
leak rate       & $\alpha$                & 0       \\ \hline
\hline
\end{tabular}
\caption{Hyperparameters used to continue the time series of the Ikeda delay system.}
\label{tab:ikeda}
\end{table}

In Fig. \ref{fig:ikedaBif} f), we show the results of the dESN trained on an Ikeda system with $\tau=100$ and compare it to the original attractor in Fig. \ref{fig:ikedaBif} a). We find that the dESN properly reproduces the chaotic attractor. Similarly as for the Mackey-Glass system, we now use the dESN trained on the Ikeda system with delay $\tau=100$ to infer the entire bifurcation diagram. In Fig. \ref{fig:ikedaBif}, we show a comparison of the original bifurcation diagram in e) and the one predicted by the dESN in j). The dESN is able to predict stable fixed points and limit cycles in the short delay range as well as the route to chaos and the chaotic attractors for longer delays. Furthermore, it precisely infers intermittent limit cycles in the chaotic regime for delays $\tau=23-25$. The inferred bifurcation diagram is again very similar; we even reproduce the fine details in the bifurcation diagram. Nevertheless, we observe small deviations between the inferred and the original attractor. Similarly, as for the Mackey-Glass system, these deviations are caused by the appearance of multistabilities in the range of $\tau,D=50-55$. Depending on the initial conditions of the dESN and Ikeda system, respectively, they either evolve on a limit cycle or on a chaotic attractor. 

\section*{ACKNOWLEDGEMENTS}
We thank André Röhm for helpful discussions.
Authors acknowledge the support of the Spanish State Research Agency, through the Severo Ochoa and María de Maeztu Program for Centers and Units of Excellence in R\&D (MDM-2017-0711 funded by MCIN/AEI/10.13039/501100011033) and through the QUARESC project (PID2019-109094GB-C21 and -C22/ AEI / 10.13039/501100011033) and the DECAPH project (PID2019-111537GB-C21 and -C22/ AEI / 10.13039/501100011033). 
MG acknowledges financial support by the European Union’s Horizon 2020 research and innovation program under the Marie Skłodowska-Curie grant agreement No. 860360 (POST DIGITAL). 
This work has been partially funded by the European Union's Horizon 2020 research and innovation programme under grant agreement No. 899265 (ADOPD).

\end{document}